\documentclass{article}
\usepackage{datarobot} % Custom DataRobot Research style

\usepackage{amsmath}
\usepackage{amssymb}
\usepackage{xspace}
\usepackage{fontawesome5}   % For github icon
\usepackage{tikz}           % For encircle
\usepackage{enumitem}       % For list formatting
\usepackage{authblk}        % For author block

\usetikzlibrary{positioning, arrows.meta, shapes.symbols, calc, decorations.pathmorphing}

\setlist[itemize]{noitemsep, topsep=0pt}
\setlist[enumerate]{noitemsep, topsep=0pt}

\newif\ifdraftmode
\draftmodefalse

\newcommand{\jointfm}{JointFM\xspace}

\newcommand{\forceditalic}[1]{{\addfontfeatures{FakeSlant=0.25}#1}}

\title{\jointfm-0.1: A Foundation Model for Multi-Target Joint Distributional Prediction}

\author{Stefan Hackmann \;|\; stefan.hackmann@datarobot.com}

\begin{document}

\maketitle

\begin{abstract}
	\noindent Despite the rapid advancements in Artificial Intelligence (AI), Stochastic Differential Equations (SDEs) remain the gold-standard formalism for modeling systems under uncertainty. However, applying SDEs in practice is fraught with challenges: modeling risk is high, calibration is often brittle, and high-fidelity simulations are computationally expensive. This technical report introduces \jointfm, a foundation model that inverts this paradigm. Instead of fitting SDEs to data, we sample an infinite stream of synthetic SDEs to train a generic model to predict future joint probability distributions directly. This approach establishes \jointfm as the first foundation model for distributional predictions of coupled time series\,---\,requiring no task-specific calibration or finetuning. Despite operating in a purely zero-shot setting, \jointfm reduces the energy loss by 21.1\% relative to the strongest baseline when recovering oracle joint distributions generated by unseen synthetic SDEs.
\end{abstract}

\section{Introduction}

\begin{figure}[h]
	\centering
	\includegraphics[width=0.8\linewidth]{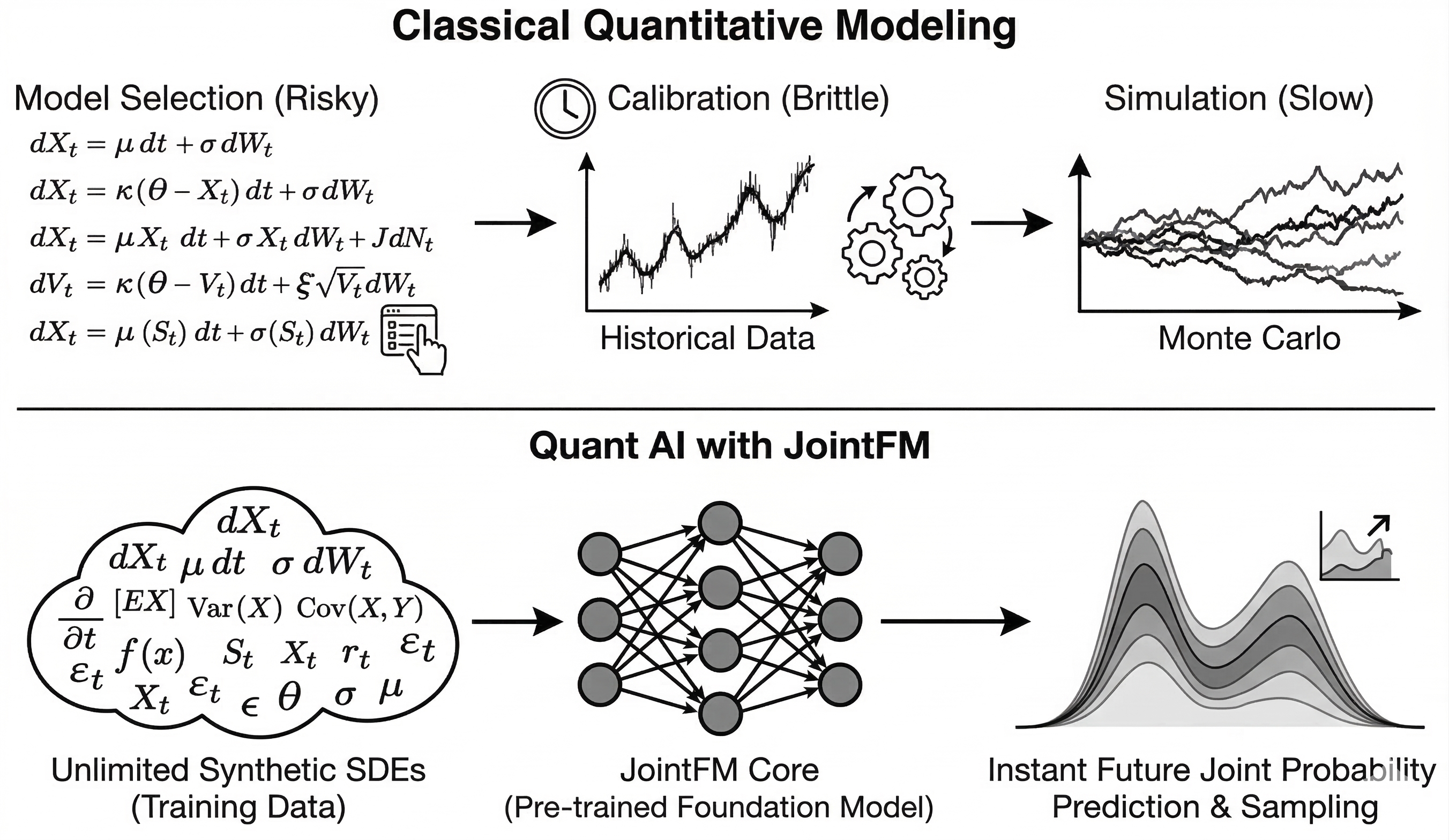}
	\caption{\textbf{\jointfm, the digital quant.} Traditional quantitative modeling follows a three-stage select-calibrate-simulate pipeline (top): choose a system of stochastic processes, fit parameters to historical data, and then simulate future paths~\cite{kloeden1992numerical,glasserman2003monte}. \jointfm replaces this workflow by pretraining on a universe of synthetic SDE dynamics and directly predicting future joint probability distributions from context in a single forward pass (bottom). Image generated with Gemini.}
	\label{fig:overview}
\end{figure}

\textbf{The Problem:} Quantitative modeling is traditionally a fragmented pipeline: first select a stochastic pro\-cess, then calibrate parameters to historical data, and finally simulate future paths. Each stage forces difficult tradeoffs: simple models such as Geometric Brownian Motion (GBM)~\cite{osborne1959brownian} miss stochastic volatility, jumps, and regime shifts, while more realistic specifications---combining correlated volatility, jumps, and regime switching---dramatically increase both calibration complexity and simulation cost. This select-calibrate-simulate pipeline is manual, brittle, often slow, and prone to overfitting. A single new data point can invalidate the entire model, requiring re-calibration, and the computational cost of simulation makes real-time risk assessment infeasible. In light of these limitations, we ask ourselves a critical question: \forceditalic{How can we make forecasting capabilities available in the age of AI that require instant responses, work with unseen data out-of-the-box, and generate forecasts as good as state-of-the-art quantitative modeling?}

\textbf{The Solution:} A ``Distributional Foundation Model'' that predicts the full joint distribution of future outcomes directly from context (zero-shot) whenever a risk decision needs to be made\footnote{We use the term ``foundation model'' not in the classical sense of a model that serves as a foundation but must be adapted to a specific task~\cite{bommasani2021opportunitiesrisksfoundationmodels} but for a model that serves as a generic, pretrained estimator for a vast collection of tasks. However, having a foundation model does not preclude further task-specific finetuning if desired.}, for instance, it becomes feasible to access the risk of a fund, like an ETF, based on its constituents on-the-fly, without needing to fit a complex multivariate SDE.

\textbf{Key Contributions:}
\begin{itemize}
	\item \textbf{Synthetic Physics Pretraining:} Prior work has explored synthetic pretraining for zero-shot fore\-cas\-ting---TempoPFN~\cite{moroshan2025tempopfn} trains on a single univariate regime-switching Ornstein--Uhlenbeck SDE~\cite{uhlenbeck1930theory}. We scale this principle to a curriculum-controlled, infinite stream of diverse multivariate SDE systems---spanning correlated diffusions, jump processes, regime switching, and non-Markovian memory\linebreak (though memory dynamics are not yet included in our training experiments).
	\item \textbf{Multi-Target Joint Distributional Predictions:} Whereas prior work predicts single future paths, univariate quantiles, or independent marginals, \jointfm is---to our knowledge---the first foundation model to train on and predict the full joint probability distribution over all target variables, explicitly capturing the dependency structure between coupled time series. This is critical for domains such as portfolio optimization, probabilistic grid balancing, and inventory management, where decisions depend on correlations between targets and optimizing them in isolation leads to suboptimal outcomes.
	\item \textbf{Unified Inference (The ``Simulation Shortcut''):} \jointfm bypasses the traditional select-calibrate-simulate pipeline. By implicitly performing model selection and calibration within its internal activations, it generates high-quality distributional predictions in a single forward pass. Because inference cost is independent of the complexity of the underlying dynamics, \jointfm eliminates the speed-quality tradeoff inherent in the traditional select-calibrate-simulate pipeline. In practice, \jointfm generates $10{,}000$ samples for $10$ targets across all $63$ horizons in about $10\,\text{ms}$ on an NVIDIA H100 GPU~\cite{jointfm_blog_portfolio}.
\end{itemize}

\section{Related Work}

\subsection{Time-Series Foundation Models (TSFMs)}
Recent advances in Time-Series Foundation Models (e.g., Chronos-2~\cite{ansari2025chronos2}, Moirai-2.0~\cite{liu2025moirai20}, TimesFM-2.5~\cite{das2023decoderonly}) have demonstrated impressive zero-shot forecasting capabilities by leveraging large-scale pretraining on diverse datasets. However, a critical limitation of these models is their focus on univariate forecasting. They typically process each time series independently and do not predict the full joint probability distribution across targets. This is a crucial distinction: risk-aware decision-making under uncertainty---e.g., portfolio optimization, supply-chain hedging, or safety-critical control~\cite{mesbah2016stochastic}---requires coherent probabilistic scenarios that capture the cross-variable dependency structure, including tail dependencies and nonlinear co-movements. Predicting each target's distribution in isolation, even accurately, is insufficient when the correlation structure is as important as the marginal distributions~\cite{embrechts2002correlation}.

\subsection{Joint Distributional Modeling}
% Classical
\textbf{Classical Statistical Models:} Traditional approaches like multivariate GARCH (e.g., DCC-GARCH~\cite{engle2002dynamic}) or copula-based methods explicitly model correlations~\cite{sklar1959fonctions}. While statistically sound, they require rigid assumptions about the underlying distribution (e.g., normality) and are computationally expensive to calibrate, especially in high dimensions~\cite{caporin2013tenthings,acar2012beyond}. They often struggle to capture complex, nonlinear dependencies and regime shifts without manual intervention~\cite{rodriguez2007measuring}.

% NN
\textbf{Neural Approaches:} Deep generative models (VAEs, GANs, and normalizing flows) provide flexible alternatives for modeling joint distributions~\cite{kingma2013autoencodingvae,goodfellow2014generativegan,rezende2015variationalflow}. However, existing multi-target methods typically require dataset-specific training or finetuning. As a result, they are not foundation models in the strict sense, because they do not zero-shot generalize to entirely new dynamics without retraining. \jointfm addresses this gap by serving as a pretrained, universal estimator of joint distributions. While normalizing flows remain a compelling option for transformer output heads, the field in 2026 is largely focused on transformer- or diffusion-based approaches for building foundation models~\cite{vaswani2017attention,ho2020denoising}.

% SDE coefficient prediction
\textbf{Neural SDE Estimation:} A complementary line of work uses neural networks to recover the governing SDE from observed data~\cite{kidger2021neuralsde,li2020scalable}. Once the SDE is identified, one can simulate unlimited forward trajectories---an attractive property. However, this approach presupposes that the data-generating process is well described by a tractable SDE family, which limits applicability outside controlled settings. In many real-world scenarios the true dynamics are unknown, non-stationary, or too complex for a tractable SDE specification, making SDE recovery impractical. Consider, for example, jointly forecasting an extensive set of coupled macroeconomic indicators: specifying, let alone calibrating, an SDE system of sufficient fidelity is hopeless given the extreme complexity of the interactions and the limited amount of historical data available. Even if such a system could be identified, the realized dynamics would likely be highly sensitive to the estimated parameters, so that small calibration errors may compound into large distributional errors over the forecast horizon. \jointfm sidesteps this issue entirely by predicting the joint distribution of future paths directly, without requiring or recovering an explicit SDE representation. Having been trained on a vast universe of complex synthetic dynamics, it can produce a best-effort distributional approximation instantly and in a zero-shot fashion.

All three paradigms above---classical calibration pipelines, neural dataset-specific training, and SDE coefficient estimation---require either substantial task-specific historical data or strong assumptions about the generative process, in contrast to a foundation model that transfers broad pretrained priors and operates in a zero-shot setting.

\section{The \jointfm Approach: The Synthetic Physics Universe}
The core hypothesis of \jointfm is that a model trained on a sufficiently diverse universe of synthetic stochastic processes will generalize to real-world processes, which can be viewed as particular realizations of these universal laws~\cite{oksendal2003stochasticdiffeq,conttankov2004financial}.\footnote{We have already demonstrated this transfer in practice: a financially-focused variant of \jointfm was applied to real market data for instant portfolio optimization~\cite{jointfm_blog_portfolio}. The experiments in this report use a generally trained foundation model that is not specialized to any particular domain.} We construct a curriculum of complexity that exposes the model to increasingly sophisticated dynamics. Training data are generated from randomly sampled SDE systems, so the model effectively never sees the same sample twice.

Formally, for each sampled SDE specification \(\phi\), we train the model (with learnable parameters \(\theta\)) to approximate the conditional density of future paths given the observed history, \(p_\theta(X_{T+1:T+H} \mid X_{1:T}, \phi)\), as implied by a generalized Stochastic Differential Equation (SDE) of the form:

\begin{equation}\label{eqn:sde}
	dX_t = \underbrace{b(t, X_t)\,dt}_{\text{Drift}} \; + \; \underbrace{\Sigma(t, X_t)\,dW_t}_{\text{Diffusion}} \; + \; \underbrace{J(t, X_t)\,dN_t}_{\text{Jumps}} \; + \; \underbrace{M(t, \mathcal{H}_t)\,dt}_{\text{Memory}} \; + \; \underbrace{R(t, X_t, S_t)\,dt}_{\text{Regime Switching}}
\end{equation}

where $X_t$ is the multivariate state, $\mathcal{H}_t$ represents the history of the path (path-dependence), $dW_t$ is a Brownian motion, $dN_t$ is a Poisson jump process, and $S_t$ is a discrete regime state. In this version, history dependence is modeled explicitly through $M(t, \mathcal{H}_t)$ and regime effects through $R(t, X_t, S_t)$; in future versions, we will extend the formulation so that drift, diffusion, and jump terms can also depend on memory and regime. The universe of physics, organized as a curriculum of increasing complexity, is summarized in Table~\ref{tab:complexity_levels}.

\begin{table}[h]
	\centering
	\label{tab:complexity_levels}
	\begin{tabular}{c l l}
		\textbf{Level} & \textbf{New Dynamic}                                                                 & \textbf{SDE Term / Mechanism}                            \\ \hline
		0              & Drift \& deterministic trends                                                        & $b(t, X_t)\,dt$                                          \\
		1              & Diffusion, nonlinear drift, sinusoidal forcing                                       & $\Sigma(t, X_t)\,dW_t,\; g(X_t)\,dt,\; s(t)\,dt$         \\
		2              & Block correlations (intra-F, intra-Y)                                                & $\operatorname{Corr}(dW^F),\;\operatorname{Corr}(dW^Y)$  \\
		3              & Cross-block correlations                                                             & $\rho_{FY}$                                              \\
		4              & Full global correlation                                                              & $\rho_{\text{global}} \in \mathbb{R}^{(M+N)\times(M+N)}$ \\
		5              & Compound Poisson jumps~\cite{merton1976optionpricing}                                & $J(t, X_t)\,dN_t$                                        \\
		6              & Regime switching --- telegraph~\cite{goldstein1951diffusion,hamilton1989newapproach} & $R(S_t)\,dt$                                             \\
		7              & Regime switching --- logistic~\cite{lando1998coxprocesses}                           & $R(X_t, S_t)\,dt$                                        \\
		8              & Regime switching --- exponential hazard~\cite{duffie2000transform}                   & $R(X_t, S_t)\,dt$                                        \\
		9              & Volterra memory~\cite{gatheral2018volatility}                                        & $M(t, \mathcal{H}_t)\,dt$                                \\
	\end{tabular}
	\caption{\textbf{Complexity curriculum}: each level introduces a new dynamic on top of all previous levels. The right column names the corresponding SDE term or mechanism: $dW^F$ and $dW^Y$ denote the Brownian increments of the $M$ exogenous features~$F$ and the $N$ target series~$Y$, respectively; $\rho_{FY}$ is the cross-block correlation between features and targets; and $\rho_{\text{global}} \in \mathbb{R}^{(M+N)\times(M+N)}$ is the full joint correlation matrix. Results in this paper use levels 0--7, only well-behaved Markovian dynamics. Future work will explore non-Markovian memory and more complex regime structures.}
\end{table}

\subsection{Industry-Agnostic Generalization}
By stochastically mixing these components (Drift, Diffusion, Jumps, Memory, Regimes), we generate a training set that encompasses the mathematical structures found across widely different sectors. A specific realization of our SDE system might resemble a regime-switching jump-diffusion (Finance), while another might resemble a Schwartz-Smith model (Commodities)~\cite{schwartz1997stochastic}. This diversity ensures that \jointfm serves as a universal foundation model, capable of adapting out-of-the-box to the specific dynamics of various industries (Table \ref{tab:sde_industry}).

\begin{table}[h]
	\centering
	\label{tab:sde_industry}
	\resizebox{\textwidth}{!}{%
		\begin{tabular}{l|llll}
			\textbf{Sector}      & \textbf{Drift / Trend ($dt$)}                                                                                         & \textbf{Diffusion / Volatility ($dW_t$)}                                                                                    & \textbf{Jumps ($dN_t$)}                                                                                       & \textbf{Memory / Regimes}                                                                                                   \\ \hline
			\textbf{Finance}     & \begin{tabular}[c]{@{}l@{}}\textbf{Vasicek / Hull--White}\\ Mean-reversion in rates\\ \& spreads.\end{tabular}        & \begin{tabular}[c]{@{}l@{}}\textbf{Correlated GBM}\\ State-dependent volatility\\ with cross-asset dependence.\end{tabular} & \begin{tabular}[c]{@{}l@{}}\textbf{Merton / Bates}\\ Price gaps upon macro\\ announcements.\end{tabular}      & \begin{tabular}[c]{@{}l@{}}\textbf{Regime Switching}\\ Bull/bear regimes driving\\ volatility \& drift shifts.\end{tabular} \\ \hline
			\textbf{Energy}      & \begin{tabular}[c]{@{}l@{}}\textbf{Ornstein--Uhlenbeck}\\ Mean-reverting demand\\ \& net load baselines.\end{tabular} & \begin{tabular}[c]{@{}l@{}}\textbf{Inhomogeneous Diffusion}\\ Intermittent wind/solar\\ supply variance.\end{tabular}       & \begin{tabular}[c]{@{}l@{}}\textbf{Spike Models}\\ Forced generator outages\\ \& extreme prices.\end{tabular} & \begin{tabular}[c]{@{}l@{}}\textbf{Regime-Switching Vol}\\ Stable vs.\ stressed regimes\\ in grid operations.\end{tabular}  \\ \hline
			\textbf{Commodities} & \begin{tabular}[c]{@{}l@{}}\textbf{Schwartz-Smith}\\ Short-term \& long-term\\ factor dynamics.\end{tabular}          & \begin{tabular}[c]{@{}l@{}}\textbf{GBM / Local Vol}\\ Spot price volatility\\ in trading strategies.\end{tabular}           & \begin{tabular}[c]{@{}l@{}}\textbf{Compound Poisson}\\ Geopolitical shocks to\\ supply chains.\end{tabular}   & \begin{tabular}[c]{@{}l@{}}\textbf{Threshold Diffusion}\\ Inventory-driven\\ price regimes.\end{tabular}                    \\ \hline
			\textbf{Logistics}   & \begin{tabular}[c]{@{}l@{}}\textbf{Geometric OU}\\ Freight rate \& transport\\ cost dynamics.\end{tabular}            & \begin{tabular}[c]{@{}l@{}}\textbf{Stochastic Lead-Time}\\ Variance in delivery\\ windows.\end{tabular}                     & \begin{tabular}[c]{@{}l@{}}\textbf{Poisson Shocks}\\ Port closures or\\ route blockages.\end{tabular}         & \begin{tabular}[c]{@{}l@{}}\textbf{Regime Switching}\\ Peak vs.\ off-peak\\ demand regimes.\end{tabular}                    \\ \hline
			\textbf{Retail}      & \begin{tabular}[c]{@{}l@{}}\textbf{Periodic OU}\\ Weekly/yearly demand\\ \& footfall cycles.\end{tabular}             & \begin{tabular}[c]{@{}l@{}}\textbf{State-Dep.~Diffusion}\\ Promotion-driven\\ demand variance.\end{tabular}                 & \begin{tabular}[c]{@{}l@{}}\textbf{Compound Poisson}\\ Viral trends \&\\ stock-out cascades.\end{tabular}     & \begin{tabular}[c]{@{}l@{}}\textbf{Regime Switching}\\ Seasonal vs.\ non-seasonal\\ buying regimes.\end{tabular}            \\
		\end{tabular}%
	}
	\caption{\textbf{Industry-Agnostic Generalization:} Mapping SDE components to industry-specific dynamics.}
\end{table}

\subsection{Procedural Data Generation Workflow}
Unlike traditional supervised learning on static datasets, \jointfm is trained on an ``infinite'' stream of procedural data. For each training batch, the workflow is as follows (see Figure~\ref{fig:training_data} in the Appendix for an example):

\begin{enumerate}
	\item \textbf{System Sampling:} We first sample a random system of SDEs from the curriculum described above. This defines the ``laws of physics'' for the current sample, including specific drift parameters, volatility surfaces, and correlation structures.

	\item \textbf{History Realization ($X_{1:T}$):} We simulate a single path of the process to serve as the context window. This history contains $D = M + N$ dimensions, where $M$ are ``feature'' covariates and $N$ are the target assets. This forms the prompt for the model.

	\item \textbf{Ground-Truth Future Distribution ($X_{T+1:T+H}$):} Because we possess the generative model, we are not limited to a single future realization. Instead, we can simulate tens of thousands of parallel future paths branching from the exact same end-state of the history. These paths approximate the true conditional joint distribution $P(X_{future} | X_{history})$.
\end{enumerate}

\subsection{Architecture Fundamentals \& Distributional Heads}
\jointfm employs a Transformer-based architecture with Variable-Factored Attention to handle a dynamic number of assets~\cite{vaswani2017attention}. The model utilizes a long context window to infer the current ``market regime'' (volatility, correlation) without explicit parameter fitting.

Crucially, the model outputs parameters for a flexible joint distribution. We compare two distinct output heads:
\begin{itemize}
	\item \textbf{Multivariate Gaussian Mixture (MV-GMM):} Predicts weights, mean vectors, and covariance matrices (via Cholesky factors) for a flexible mixture of Gaussians~\cite{mclachlan2000finitemixture}.
	\item \textbf{Skewed Student-t Mixture:} Predicts weights, degrees of freedom ($\nu$), location ($\mu$), scale matrix ($\Sigma$), and skewness ($\alpha$) for a mixture of multivariate skewed Student-t components. This parametric form explicitly captures heavy tails and asymmetry, essential for financial risk modeling~\cite{azzalini2003distributions}.
\end{itemize}

\section{Experiment: Zero-Shot Synthetic Distribution Recovery}
\textbf{Concept:} A fundamental test of a foundation model is its ability to recover the ground-truth data\--gen\-er\-at\-ing process from unseen synthetic data. We evaluate \jointfm's zero-shot capability to accurately model complex, multivariate distributions without finetuning.

\textbf{Setup:} We draw 1{,}000 unseen SDE systems at curriculum level~7 (the highest Markovian level, which includes all dynamics up to logistic regime switching, including jumps).
Each system comprises $N = 10$ coupled target series with no exogenous features ($M = 0$).
From each sampled SDE we simulate a single history realization of $T = 504$ input steps (spanning a simulated time window of $t_{\text{in}} = 2.0$), which serves as the model's context.
For the ground truth, we branch 1{,}000 independent future paths of $H = 63$ steps ($t_{\text{out}} = 0.25$) from the terminal state of the history, yielding a Monte Carlo approximation to the true conditional joint distribution $P(X_{T+1:T+H} \mid X_{1:T})$.
All models produce 1{,}000 forecast samples, and metrics are computed independently at each of the 63~forecast horizons.
\jointfm is evaluated in pure zero-shot mode (no finetuning on the test instances); baselines are fitted directly to the observed history window.

\textbf{Metrics:} We employ three distribution-matching metrics:
\begin{itemize}
	\item \textbf{Energy Loss:} Measures the statistical distance between generated and ground-truth joint samples, capturing global distribution mismatch. This is the primary metric for this work, as it is a proper multivariate scoring rule that fully evaluates the joint distribution---including all pairwise and higher-order dependencies.
	\item \textbf{Marginal Energy:} Applies the energy-distance objective per target variable independently, verifying marginal calibration. Because it evaluates each variable in isolation, it is blind to the correlation structure that is central to this work.
	\item \textbf{CRPS-sum (Continuous Ranked Probability Score on Sums):} A proper scoring rule that assesses calibration and sharpness of the distribution of the aggregated target $\sum_i X_i$~\cite{gneiting2007strictlyproper}. While indirectly sensitive to correlations via the variance of the sum–but without being able to distinguish different correlation structures that yield the same aggregate variance.
\end{itemize}
Of these three metrics, only Energy Loss directly evaluates the full joint distribution. Marginal Energy and CRPS-sum are included for completeness and to demonstrate that \jointfm's joint modeling does not come at the expense of marginal calibration, but they are of secondary relevance to our core contribution.

\textbf{Comparison:}
\begin{enumerate}
	\item \textbf{\jointfm (Ours):} Zero-shot prediction using both the Skewed Student-t and MV-GMM heads.
	\item \textbf{DCC-GARCH:} A classic multivariate GARCH model with dynamic conditional correlations, fitted to the history~\cite{engle2002dynamic}. We use the open-source implementation from \url{https://github.com/srivastavaprashant/mgarch} with a default multivariate normal distribution and patch \texttt{scipy.optimize.minimize} to enforce\linebreak\texttt{maxiter=500}.
	\item \textbf{Historical Simulation:} Non-parametric bootstrap resampling from the historical window~\cite{hendricks1996evaluation}.
\end{enumerate}

We restrict baselines to methods that can be applied directly to the observed history window without prior training: DCC-GARCH (parametric fit) and Historical Simulation (non-parametric bootstrap). A controlled comparison with supervised neural baselines trained on comparable data budgets is left for future work.

\begin{figure}[t]
	\centering
	\includegraphics[width=\linewidth]{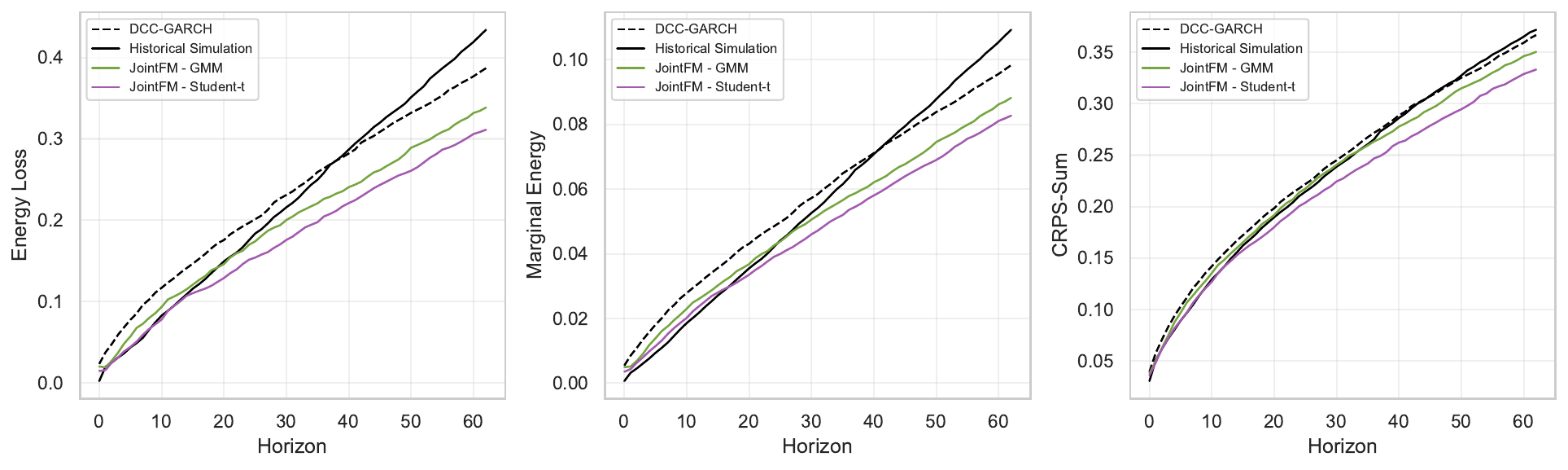}
	\caption{\textbf{Recovering joint distributions.} Zero-shot synthetic distribution recovery across model families and baselines. Lower values indicate better distributional match on energy loss, marginal energy, and CRPS-sum. \jointfm variants consistently outperform Historical Simulation and DCC-GARCH, with the performance gap becoming increasingly pronounced as the forecast horizon lengthens.}
	\label{fig:synthetic_recovery}
\end{figure}

\textbf{Results (Figure~\ref{fig:synthetic_recovery}):} We observe that \jointfm consistently outperforms the baselines across all three metrics. When predicting 63 steps into the future, JointFM - Student-t achieves 19.6\% lower energy loss, 15.9\% lower marginal energy loss, and 9.1\% lower CRPS-sum relative to the best-performing baseline. Averaged across all forecast horizons, JointFM - Student-t achieves 21.1\% lower energy loss, 15.8\% lower marginal energy loss, and 7.3\% lower CRPS-sum relative to the best-performing baseline.

Among the baselines, Historical Simulation is competitive at short horizons where recent returns remain representative of the near-term future, but its performance degrades markedly as the forecast horizon lengthens. DCC-GARCH shows a rather disappointing performance overall; however, this is largely attributable to the limited context of $T = 504$ steps, which sits at the lower end of the sample sizes typically required for reliable GARCH-family estimation~\cite{engle2002dynamic}.

\section{Limitations \& Future Work}

\textbf{Dynamics Coverage.}
\jointfm can only recover dynamics that are represented in its training curriculum. The current version (levels 0--7) covers well-behaved Markovian processes with constant or state-scaled volatility, correlated diffusions, Compound Poisson jumps, and regime switching. Levels~8 (exponential hazard regime switching) and~9 (Volterra memory and rough-path effects) are implemented but excluded from our experiments because they pose a significantly higher level of complexity. Beyond these existing levels, we plan to extend the curriculum to richer volatility specifications---including stochastic volatility (e.g., Heston, SABR), local-stochastic volatility, and rough volatility models---which are widely used in quantitative finance but currently outside the sampler's scope.

\textbf{Proof-of-Concept Scale.}
\jointfm-0.1 is a scientifically evaluated proof-of-concept trained with a context window of $T = 504$ input steps, $M = 0$ exogenous features, up to $N = 10$ target series, and a forecast horizon of $H = 63$ steps. We intend to scale both the data pipeline and the model, and are investigating architectural changes that will make scaling to longer contexts, higher-dimensional target spaces, and richer feature sets practical. Early results indicate that scaling beyond 100 features and 100 targets is feasible but in combination with an increased context length and number of output horizons, hardware constraints become a bottleneck.

\textbf{Observed Dynamics Fallacy.}
A fundamental limitation of any purely data-driven approach is that if a specific dynamic (e.g., a jump process representing default risk) has not yet occurred within the context window, the model may fail to assign probability mass to it, whereas a human practitioner would incorporate such structural risks a priori. To bridge this gap, future iterations will incorporate explicit time-series labeling during training. Because we control the data-generating process, we can supply categorical tokens (e.g., ``Jump-Diffusion'', ``Mean-Reverting'') alongside the time-series data, enabling a ``hybrid inference'' mode where practitioners optionally guide the model's priors with domain knowledge while retaining purely data-driven inference for novel situations.

\textbf{Data Types.}
\jointfm currently operates exclusively on continuous (float) time series. Many real-world forecasting problems, however, involve categorical, ordinal, or mixed-type variables (e.g., credit ratings, demand categories, or discrete event indicators). Extending the model to handle heterogeneous data types is a natural next step and will require adapting both the synthetic data generation pipeline and the distributional output heads.

\textbf{SDE Sampling Infrastructure.}
The current SDE sampling pipeline relies on pre-configured dynamics families. We aim to build a more general, compositional SDE sampler that supports targeted training on under-represented regimes and enables systematic exploration of the vast space of stochastic systems---re\-mov\-ing the dependency on hand-designed curriculum entries.

\section{Conclusion}
We introduced \jointfm, the first foundation model capable of predicting future joint probability distributions over coupled time series. By replacing the traditional select-calibrate-simulate pipeline with a single forward pass, \jointfm eliminates the need for task-specific calibration or finetuning and delivers full distributional predictions---including cross-variable dependencies, tail behavior, and asymmetries---instantly.

Our experiments confirm that this approach works: operating in a purely zero-shot setting, \jointfm successfully recovers oracle joint distributions generated by unseen synthetic SDE systems, consistently outperforming classical baselines across all evaluated metrics. This constitutes, to our knowledge, the first empirical demonstration that a pretrained neural model can generalize across the space of multivariate stochastic processes without access to the underlying equations or any retraining.

This capability opens the door for generative AI to serve business domains that have historically been the exclusive province of quantitative modeling---portfolio risk assessment, energy market dispatch, supply-chain hedging, and any decision problem where the joint distribution of future outcomes governs the optimal action. Where these tasks previously demanded bespoke SDE specifications, manual calibration, and computationally expensive Monte Carlo simulations, \jointfm provides an instant, general-purpose alternative whose cost is independent of the complexity of the underlying dynamics.

Because predictions are generated in a single forward pass at negligible latency, \jointfm is natively compatible with agentic AI workflows that require real-time probabilistic reasoning. We expect models of this kind to become the quantitative backbone of the agentic age---enabling autonomous systems to make risk-aware decisions on the fly, without waiting for a human-in-the-loop calibration cycle.

These results mark the beginning. The current proof-of-concept covers well-behaved Markovian dynamics; scaling to richer physics, longer contexts, and higher-dimensional target spaces is the natural next step. We believe the synthetic-physics pretraining paradigm will extend to any domain whose dynamics can be expressed as stochastic differential equations, establishing a clear path toward real-time, universal quantitative modeling across industries---from finance and energy to commodities, logistics, and retail.

\section*{Acknowledgments}
The author gratefully acknowledges Alexander Conway, Matthew Hausknecht, and Mark L.\ Steadman for valuable discussions throughout the course of this work. Their expertise in generative modeling and classical time-series analysis provided helpful perspective during the development of this work.

\bibliographystyle{plain}
\bibliography{references}

\clearpage
\appendix
\section{Training Data}
\begin{figure}[h]
	\centering
	\includegraphics[width=\linewidth]{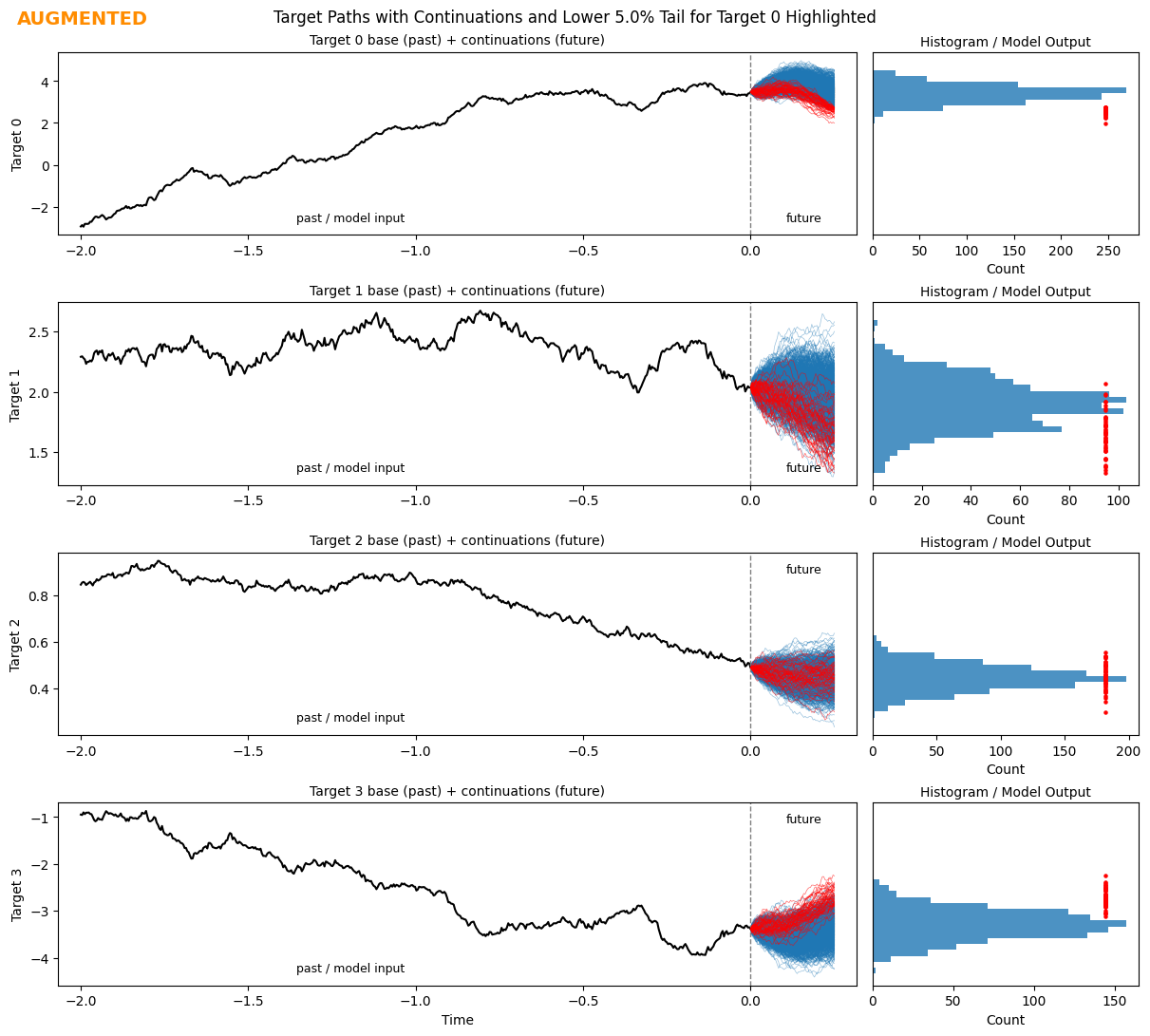}
	\caption{\textbf{Training data.} This example stems from SDE sampler as it is configured for our recovery experiment but showing only four targets.}
	\label{fig:training_data}
\end{figure}

\end{document}